\documentclass[10pt,twocolumn,letterpaper]{article}

\usepackage[pagenumbers]{cvpr}

\usepackage{graphicx}
\usepackage{cite}
\usepackage{amsmath,amssymb,amsfonts}
\usepackage{algorithmic}
\usepackage{graphicx}
\usepackage{textcomp}
\usepackage{xcolor}
\usepackage{color}
\usepackage{colortbl}
\usepackage{xspace}
\usepackage[labelfont=bf]{caption}
\usepackage{bbding}
\usepackage{multirow}
\usepackage{booktabs,array}
\usepackage[utf8]{inputenc} 

\newcommand{\bb}[1]{\textbf{#1}}

\newcommand{\mv}{\vspace{-2mm}}
\newcommand{\point}{\item \mv}

\newcolumntype{C}[1]{>{\centering\arraybackslash}p{#1}}

\makeatletter
\newcommand\thefontsize[0]{{The current font size is: \f@size pt\par}}
\makeatother

\usepackage[pagebackref,breaklinks,colorlinks]{hyperref}

\usepackage[capitalize]{cleveref}
\crefname{section}{Sec.}{Secs.}
\Crefname{section}{Section}{Sections}
\Crefname{table}{Table}{Tables}
\crefname{table}{Tab.}{Tabs.}

\begin{document}

\title{Intriguing properties of synthetic images: \\ from generative adversarial networks to diffusion models}

\author{
Riccardo Corvi\textsuperscript{1} \ \ \ 
Davide Cozzolino\textsuperscript{1} \ \ \ 
Giovanni Poggi\textsuperscript{1} \ \ \ 
Koki Nagano\textsuperscript{2} \ \ \ 
Luisa Verdoliva\textsuperscript{1} \\[2mm]
{\textsuperscript{1}University Federico II of Naples \ \ \ \ \ \textsuperscript{2} NVIDIA}}

\twocolumn[{%
	\renewcommand\twocolumn[1][]{#1}%
	\maketitle
	\begin{center}
	    \vspace{-0.5cm}
            \includegraphics[width=0.99\linewidth, trim=0 0 0 0, clip, page=1]{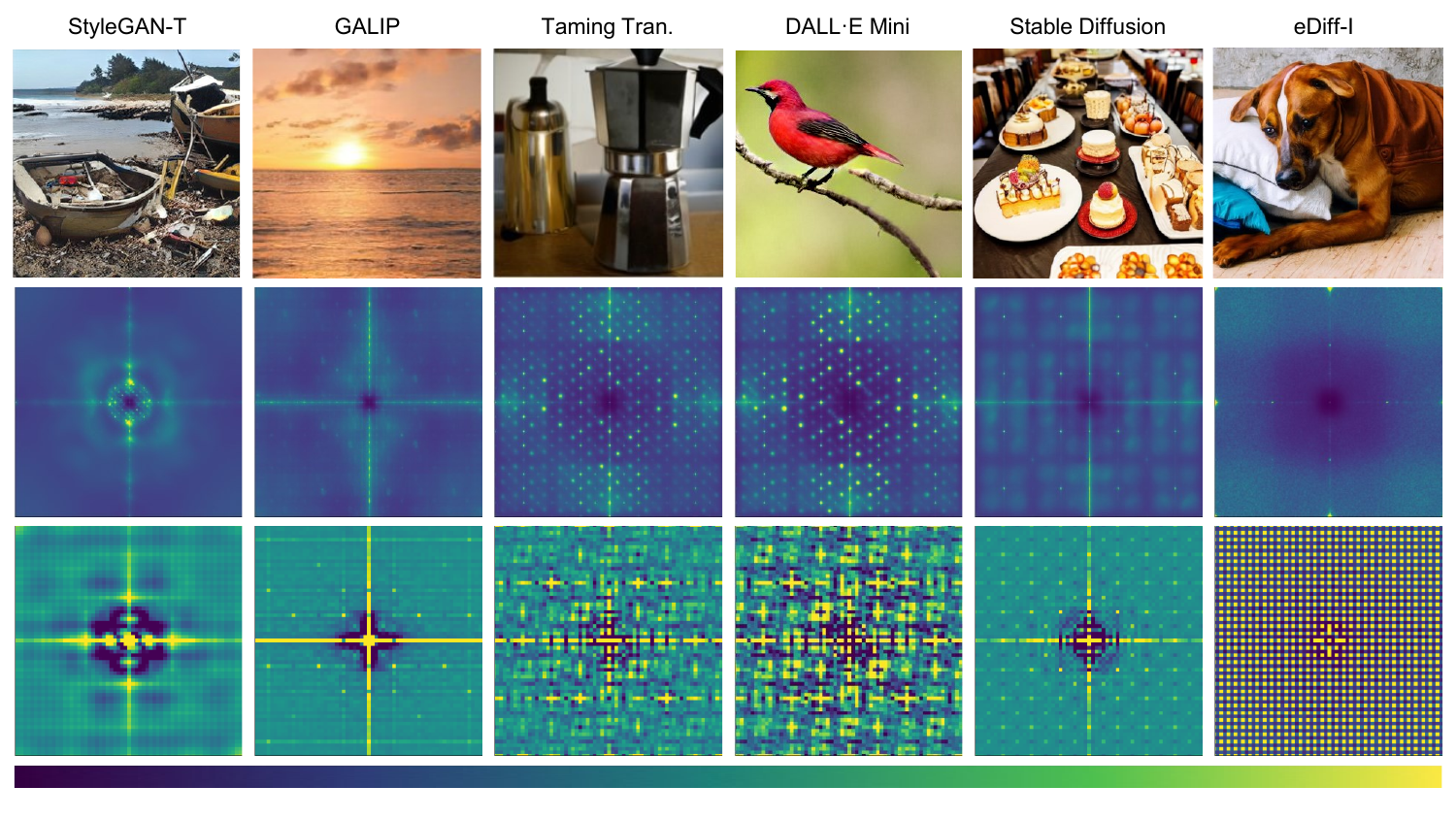}
		\vspace{-0.2cm}
		\captionof{figure}{
		Examples of synthetic images (top) generated by state-of-the-art models:
StyleGAN-T \cite{sauer2023stylegan}, GALIP \cite{tao2023galip}, Taming Transformers \cite{esser2021taming},
DALL$\cdot$ E Mini \cite{Dayma_DALLE_Mini_2021}, Stable Diffusion \cite{stablediffusion2022}, eDiff-I \cite{balaji2022ediffi}.
Each model leaves peculiar traces in the generated images,
which can be regarded as a sort of artificial fingerprint and used for forensic analyses.
They are typically visible in the frequency domain as spectral peaks in the power spectra (middle),
or in the spatial domain as anomalous regular patterns in the autocorrelation (bottom). 
It is easy to observe that models based on the same architecture, like Taming Transformers and DALL$\cdot$ E Mini, give rise to similar artifacts.
} \label{fig:teaser}
	\end{center}
}]

\begin{abstract}
\vspace{-0.5cm}
Detecting fake images is becoming a major goal of computer vision.
This need is becoming more and more pressing with the continuous improvement of synthesis methods based on Generative Adversarial Networks (GAN),
and even more with the appearance of powerful methods based on Diffusion Models (DM).
Towards this end, it is important to gain insight into which image features better discriminate fake images from real ones.
In this paper we report on our systematic study of a large number of image generators of different families,
aimed at discovering the most forensically relevant characteristics of real and generated images.
Our experiments provide a number of interesting observations and shed light on some intriguing properties of synthetic images:
(1) not only the GAN models but also the DM and VQ-GAN (Vector Quantized Generative Adversarial Networks) models give rise to visible artifacts in the Fourier domain and exhibit anomalous regular patterns in the autocorrelation;
(2) when the dataset used to train the model lacks sufficient variety, its biases can be transferred to the generated images;
(3) synthetic and real images exhibit significant differences in the mid-high frequency signal content, observable in their radial and angular spectral power distributions.
\end{abstract}

\begin{figure*}
        \centering
        \includegraphics[width=0.95\linewidth, trim=0 5 0 0, clip, page=2]{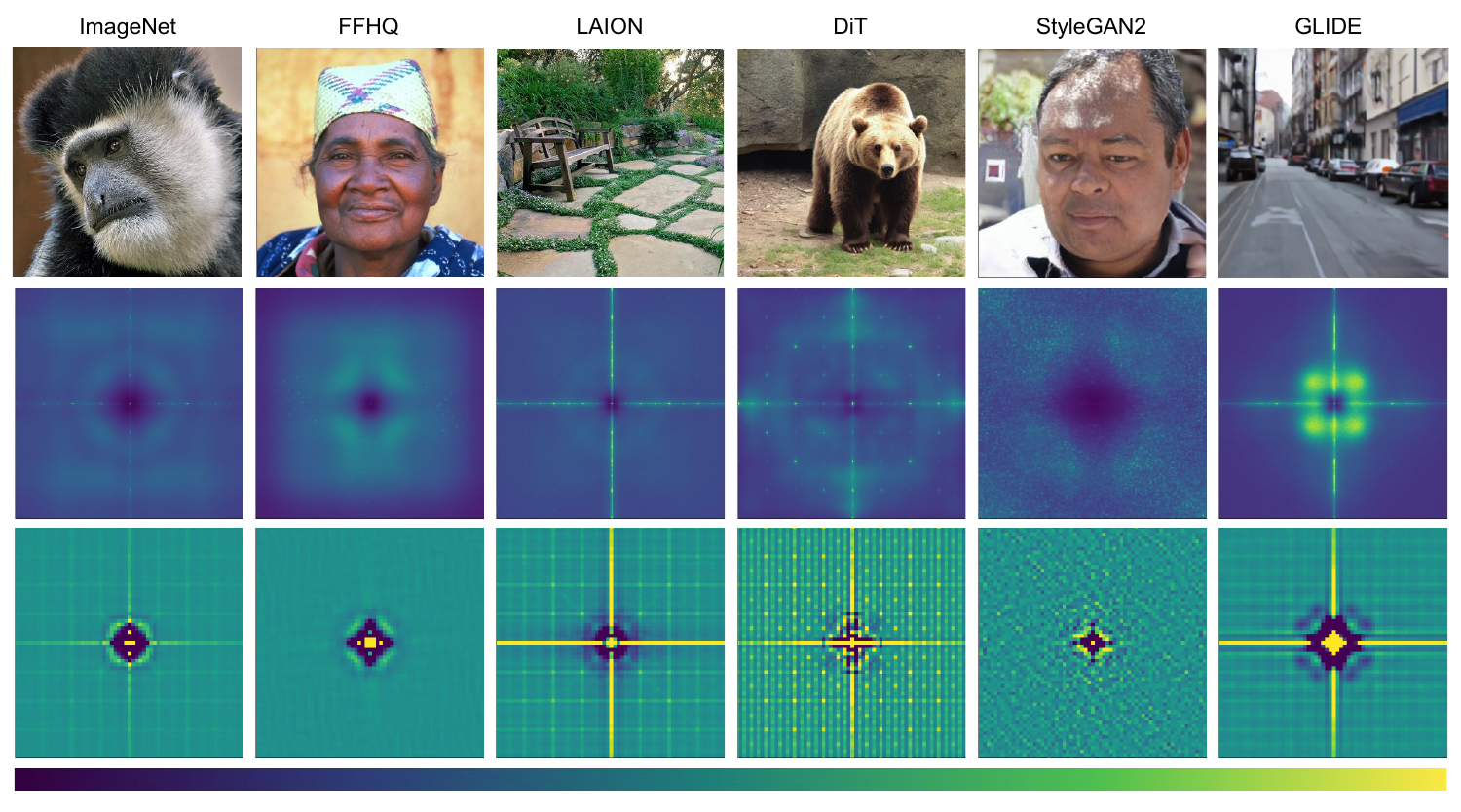}

\caption{Examples of real and synthetic images drawn from popular datasets. 
Real datasets (left): Imagenet \cite{deng2009imagenet}, FFHQ \cite{karras2019style}, LAION \cite{schuhmann2021laion}. Synthetic datasets (right): DiT \cite{peebles2022scalable}, StyleGAN2 \cite{karras2020analyzing}, GLIDE \cite{nichol2021glide}.
For each dataset, we also show the average power spectra (middle) and autocorrelation (bottom) of noise residuals. 
It is possible to observe that real images embed some traces of JPEG artifacts (see also Section 4.2) while all generated images exhibit anomalous periodic patterns in the spatial domain.}
	\label{fig:second_teaser}
\end{figure*}

\section{Introduction}
\label{sec:intro}

In recent years, deep learning-based generated images have gained immense popularity in industries such as entertainment, advertising, and design.
These are high-quality images of impressive photorealism \cite{Lago2021more, Nightingale2022AI}.
Especially the recent text-image generators, which rely on text descriptions or preliminary sketches as a guide,
can allow image creation across countless categories \cite{Dayma_DALLE_Mini_2021, nichol2021glide,ramesh2022hierarchical,balaji2022ediffi,tao2023galip,sauer2023stylegan,kang2023scaling}.
While this technology offers tremendous value to artists, game designers, and creative individuals, it also comes with risks.
The potential security and ethical issues that can arise from the misuse of these images cannot be ignored.

Research on synthetic image detection has received a huge boost in recent years \cite{Marra2018, McCloskey2019detecting},
especially since the first realistic images generated by Generative Adversarial Networks \cite{Zhu2017unpaired, karras2018progressive} appeared.
Some works focus mainly on the inability of such generators to perfectly reproduce the high-level semantic features of natural images.
In fact, they can produce visible artifacts, such as chromatic anomalies or lack of natural symmetries.
For example, faces generated by GANs may have differently colored left and right eyes, unconvincing eye reflexes, or irregular pupil shapes \cite{Matern2019exploiting, Guo2022eyes}.
Similarly, some recent studies on images generated by diffusion models \cite{Farid2022perspective, Farid2022lighting}
pointed out that the absence of explicit 3D modeling for objects and surfaces results in asymmetries in shadows and reflections.
In addition, a global inconsistency in lighting can be observed to some extent.

However, the generation methods keep improving over time and show fewer and fewer synthesis artifacts.
Therefore, other approaches focus on low-level features that are related to the generative model architecture and can be exposed by removing the high-level semantic content of the captured scene \cite{Marra2019DoGAN, Yu2019}.
In fact, the complex processing pipeline needed to generate synthetic images
inevitably leads to the introduction of digital artifacts that characterize the specific architecture and significantly differ from those typical of traditional acquisition devices. Some works have already analyzed the traces present in the synthetic images working in the Fourier domain.
However, the analysis mainly focused on artifacts caused by GAN models,
such as those due to the up-sampling steps used to achieve the desired spatial resolution \cite{Zhang2019, Dzanic2020Fourier}.
To date, little attention has been paid to more recent and promising architectures,
such as those based on auto-encoders and diffusion models
\cite{Corvi2022detection, Sha2022fake, Ricker2022towards, Ojha2023towards}.

In this work we carry out a systematic investigation of the traces left not only by
established GAN-based generators but also by several recent VQ-based and DM-based generators.
To this end, we consider second-order statistics both in the spatial domain and in the frequency domain.
For the spatial domain, we use the average autocorrelation function
while its transform, the average power spectrum, is used for frequency-domain analyses
together with some derived quantities: the radial and angular spectra.
Examples of power spectra and (central crop) autocorrelations are shown in Fig.~\ref{fig:teaser} and Fig.~\ref{fig:second_teaser}. 

As a result of our analyses we contribute the following main observations:
\begin{itemize}
\point  To date, no generator appears to be artifact-free.
        Most of the times, artifacts are clearly spotted as strong peaks in the power spectra of noise residuals,
        but some recent methods have succeeded in reducing them.
        Even in these cases, unnatural regular patterns are typically observed in the autocorrelations of noise residuals.
\point  Artifacts ({\it e.g.}, due to compression) present in the dataset used to
        train a given model can be transferred to generated images.
        This is especially evident when the training dataset has limited variety and strong biases.
\point  Synthetic and real images may be distinguished not only by the presence of artifacts but also because of mismatched statistical features of the high-level real or generated content.
        We highlight such consistent differences by studying the radial and angular spectra of the sources, derived from the average power spectra.
\end{itemize}
In the following of the paper, we account for related work, describe the generative models under investigation,
report on experimental results, discuss their significance, and finally draw conclusions.

\begin{table*}[t!]
    \centering
    \scalebox{0.96}{
    \setlength{\tabcolsep}{3pt}
    \renewcommand\arraystretch{1.0}
       \begin{tabular}{lclccc}
       \toprule & &
  \multicolumn{1}{c}{\textbf{Method}} &
  \multicolumn{1}{c}{\textbf{Backbone}} &
  \multicolumn{1}{c}{\textbf{Dataset}} &
  \multicolumn{1}{c}{\textbf{Conditioning}} \\ \midrule
  \multirow{5}{*}{\rotatebox[origin=l]{90}{GAN}}
  & \cite{brock2018large} & BigGAN            & ResNet-style  & ImageNet     & Class \\
  & \cite{karras2020analyzing} & StyleGAN2    & StyleGAN like   & FFHQ                      & Uncond. \\
  & \cite{karras2021alias} & StyleGAN3        & StyleGAN like   & FFHQ                      & Uncond. \\
  & \cite{sauer2023stylegan} & StyleGAN-T     & StyleGAN like   & CC12M, CC, YFCC100m, Redcaps, LAION-aesthetic-6+     & Text \\
  & \cite{tao2023galip} & GALIP               & Transf. based   & CUB, MS-COCO, CC3M, CC12M                      & Text \\ \midrule
  \multirow{2}{*}{\rotatebox[origin=l]{90}{VQ}}
  & \cite{esser2021taming} & Taming Transf.   & VQGAN + Transf. & ImageNet, FFHQ          & Class, Uncond.\\
  & \cite{Dayma_DALLE_Mini_2021} & DALL·E Mini & VQGAN + Transf. &
CC, YFCC100M, CC12M & Text \\ \midrule
  \multirow{5}{*}{\rotatebox[origin=l]{90}{DM}}
  & \cite{song2021score} & Score-SDE          & U-Net       & FFHQ                  & Uncond. \\
  & \cite{dhariwal2021diffusion} & ADM        & U-Net       & ImageNet, LSUN        & Class, Uncond. \\
  & \cite{nichol2021glide} & GLIDE            & U-Net       & CC, YFCC100M, Wikipedia text-image pairs                      & Text\\
  & \cite{ramesh2022hierarchical} & DALL·E 2  & U-Net       &  Unknown              & Text\\
  & \cite{balaji2022ediffi} & Ediff-I         & Mul. U-Nets &  Unknown              & Text\\ \midrule
  \multirow{3}{*}{\rotatebox[origin=l]{90}{LDM}}
  & \cite{rombach2022high} & Latent Diff.     & VAE + U-Net & ImageNet, LSUN      & Class, Uncond.\\
  & \cite{stablediffusion2022} & Stable Diff. & VAE + U-Net & LAION      & Text\\
  & \cite{peebles2022scalable} & DiT          & VAE + ViT   & ImageNet      & Class \\
  \bottomrule
    \end{tabular}
    }
    \caption{Main properties of the generative models analyzed in this work.}
    \label{tab:ModelTables}
\end{table*}

\section{Related Work}
\label{sec:related}

In this section, we describe previous work that shows the evidence of peculiar invisible traces embedded in the synthetic images both in the spatial and in the frequency domain.

\vspace{2mm} \noindent
\bb{Artificial fingerprints.}
Early research on synthetic images has shown that,
similar to real cameras that imprint each image with a device- or camera-specific signature \cite{Lukas2006, Cozzolino2020noiseprint},
GAN architectures also embed a fingerprint in each generated image.
Such subtle traces, that are originated by the processing steps used in the image synthesis process,
betray the artificial nature of the images.
Several detectors \cite{Marra2019DoGAN, Yu2019, Asnani2021reverse} rely on these traces to expose synthetic images
and even to trace back the individual architecture that generated them.
Differently from these works, we rely on the second order statistics of the images
to better highlight the patterns created by the generation process.

\vspace{2mm} \noindent
\bb{Spectral discrepancies.}
It is well known that synthetic images generated by GANs present anomalies in the frequency domain
and cannot perfectly reproduce the spectral distribution of natural images.
More specifically, in \cite{Zhang2019} for the first time evidence of such traces was shown, which appear as peaks in the Fourier domain and can be used as discriminating features for the development of a forensic detector.
These traces are caused by the up-sampling operation present in the decoder architecture of the generation process
which causes a marked aliasing phenomenon~\cite{karras2021alias}.
Further studies \cite{Dzanic2020Fourier} showed that real and synthetic images
exhibit a different decay of the Fourier spectrum at the highest frequencies.
The inability to correctly reproduce the spectral distributions of the real data used for training
is often attributed to up-sampling methods \cite{Durall2020, Frank2020leveraging}.
Some works also propose a modification to the generator loss in order to better fit the spectral decay of natural images \cite{Durall2020, jiang2021focal}.
Interestingly, it has been recently shown that also diffusion models are affected by a frequency bias
and are not able to perfectly reproduce high frequencies and image details \cite{Yang2022diffusion}.

Based on these findings, several methods for GAN image detection
rely on the high frequency discrepancies between real and generated images.
However, it is not easy to exploit such traces in the challenging conditions of real-world applications \cite{Gragnaniello2021GAN, Chandrasegaran2021closer}.
In this work we show that also middle frequencies can play a significant role for discrimination.
In particular, we observe that the generators possess an anisotropic behaviour
and struggle to reproduce details along the diagonal directions.

\section{Generative Models}
This section aims at describing the generators of synthetic images analyzed in this work and their most important properties.
A list of the models is reported in Tab.~\ref{tab:ModelTables}.

\vspace{2mm} \noindent
\bb{Generative Adversarial Networks.}
GANs adopt a training strategy based on a min-max game between two networks: a generator, and a discriminator~\cite{Goodfellow2014GAN}.
The former tries to create realistic samples, able to fool the discriminator,
while this latter tries to correctly distinguish real from generated samples.
In 2019, Brock et al. proposed BigGAN \cite{brock2018large} a class-conditional image synthesizer.
Karras et al. proposed a series of GAN architectures for unconditional image generation, called StyleGAN \cite{karras2019style,karras2020analyzing, karras2021alias},
where the latent code controls the modulation of the convolutional kernels of the generator.
StyleGAN-T \cite{sauer2023stylegan} uses a StyleGAN-like architecture to generate text-guided images
relying on CLIP (Contrastive Language-Image Pre-Training) \cite{radford2021learning}
a powerful model that matches an image with its textual description.
Recently, Tao et al. proposed GALIP (Generative Adversarial CLIP) \cite{tao2023galip}
a GAN for text-to-image synthesis where both the discriminator and generator architectures are based on CLIP.

\vspace{2mm} \noindent
\bb{Vector Quantized Generative Networks.}
The Vector Quantized Variational Autoencoders (VQ-VAE) proposed in \cite{van2017neural}
are auto-encoders with an encoder network to compress data into a low-dimensional latent space
and a decoder network to reconstruct the original data from the latent space.
In particular, the VQ-VAEs quantize the latent space using a discrete codebook,
which allows the model to learn latent vectors with a more meaningful and structured prior.
In \cite{chen2020generative}, the prior is derived from an autoregressive model based on a transformer network,
while in \cite{esser2021taming}, the authors propose VQ-GAN,
a variant of the original VQ-VAE that induces an adversarial loss.
Both Taming Transformers \cite{esser2021taming} and DALL·E Mini \cite{Dayma_DALLE_Mini_2021}, considered in this work, are VQ-GANs with a transformer network as the autoregressive model of the prior.

\vspace{2mm} \noindent
\bb{Diffusion Models.}
To generate synthetic samples mimicking real ones, diffusion models rely on the inversion of an additive noise process.
The model takes as input a noisy image composed of white noise and image content,
and generates progressively less noisy versions of it until reaching the desired noiseless output.
Recently, DMs~\cite{ho2020denoising, sohl2015deep, song2019generative, karras2022elucidating} have gained popularity for the task of image synthesis due to their favorable properties, such as stable training and better mode coverage compared to previous methods~\cite{Goodfellow2014GAN}.
Ho et al. \cite{ho2020denoising} first demonstrated the capability of diffusion models to generate high-quality images.
Later, Dhariwal et al. \cite{dhariwal2021diffusion} showed that DMs can produce images of better quality than those generated by GANs.
The use of U-Net architecture or its variations is prevalent among diffusion models \cite{song2021score, nichol2021glide, ramesh2022hierarchical, balaji2022ediffi, rombach2022high, stablediffusion2022}.
In particular, \cite{nichol2021glide, ramesh2022hierarchical} extend the previous work \cite{dhariwal2021diffusion}
to image generation from text based on CLIP.
The recent work Ediff-I \cite{balaji2022ediffi} adopts multiple U-Net models specialized for different synthesis stages.

\vspace{2mm} \noindent
\bb{Latent Diffusion Models.}
These models, proposed in \cite{rombach2022high}, combine variational autoencoders (VAEs) with diffusion models.
a VAE is used to project images into a latent space and back.
In this low-dimensional space, a diffusion model can generate the desired latent vector with limited computational cost.
This combination allows for the generation of high-dimensional images with good quality
and with a reduced computational burden compared to other diffusion models.
A noteworthy model of this class is Stable Diffusion \cite{stablediffusion2022},
which is part of an open-source project and is trained on the 5.85 billion images of the LAION dataset \cite{schuhmann2021laion}.
A recent work, Diffusion Transformers (DiTs) \cite{peebles2022scalable},
proposes to replace the commonly-used U-Net architecture with a transformer architecture.


\section{Forensic Analysis}

A crucial step in developing increasingly effective detectors is understanding
which image features can help distinguish fake images from real ones.
Interestingly, a few carefully designed simple experiments provide us already with several insightful observations.
This section serves as a guided journey through some of these experimental results, with the aim of conveying the importance of these findings.
Our analysis is carried out by relying on second-order statistics observed
both in the spatial domain, the image autocorrelation function, and in the frequency domain, the image power spectrum\footnote{In the literature, ``power'' is sometimes replaced by ``energy'', but it is only a matter of theoretical image modeling.}.

For all sources of interest we consider a set of $I=1000$ images, either real or generated under controlled conditions.
For a single image $x_i(m,n)$ the autocorrelation function is
\newcommand{\dm}{{\Delta m}}
\newcommand{\dn}{{\Delta n}}
\begin{equation}
    R_{x_i}(\dm,\dn) = \langle x_i(m,n)x_i(m+\dm,n+\dn) \rangle
\end{equation}
where $\langle\cdot\rangle$ indicates spatial average.
The average autocorrelation of the source should be obtained through statistical averaging,
which we approximate by averaging on all $I$ images
\begin{equation}
    R_{x}(\dm,\dn) = \frac{1}{I} \sum_{i=1}^I R_{x_i}(\dm,\dn)
\end{equation}
This function summarizes the statistical correlation between pixels separated by a $(\dm,\dn)$ displacement.

For frequency domain analyses, we start from the Fourier transform of the $M \times N$ image 
\begin{equation}
\begin{split}
    X_i(k,l) &= {\cal F}[x_i(m,n)] = \\
             &= \sum_{m=1}^{M} \sum_{n=1}^{N} x_i(m,n) e^{-j2\pi(\frac{k}{M}m+\frac{l}{N}n)}
\end{split}
\end{equation}
and obtain the source power spectrum, again, by averaging all individual power spectra
\begin{equation}
       S_x(k,l) = \frac{1}{I} \sum_{i=1}^{I} S_{x_i}(k,l) = \frac{1}{I} \sum_{i=1}^{I} |X_i(k,l)|^2
\end{equation}
The power spectrum accounts for the fraction of the total image power
concentrated at a given (horizontal,vertical) frequency pair $(\frac{k}{M},\frac{l}{N})$.
Note that these two functions are themselves related by a frequency transform
\begin{equation}
    S_x(k,l) = {\cal F}[R_{x}(\dm,\dn)]
\end{equation}
hence redundant, in theory.
However, some phenomena are more visible in the spatial domain and some in frequency, so we study both of them.

Often, phenomena of interest are only visible in the noise residuals, 
obtained by removing the high-level semantic content from the image by means of a denoising filter.
\begin{equation}
    r_i(m,n) = x_i(m,n) - {\cal D}(x_i(m,n);\sigma)
\end{equation}
In our experiments, we use always the denoiser by Zhang {\it et al.} \cite{Zhang2017beyond} with noise parameter $\sigma$=1.

All autocorrelations and spectra, associated to images or residuals, are normalized to the average power per pixel computed as the mean value of $S_x(k,l)$.
Note that we show only the central $65\times 65$ crop of the autocorrelation function.

\subsection{Architecture related traces}

\begin{figure}[t!]
        \centering
	\includegraphics[width=1.0\linewidth,clip,trim=183 150 183 0,page=5]{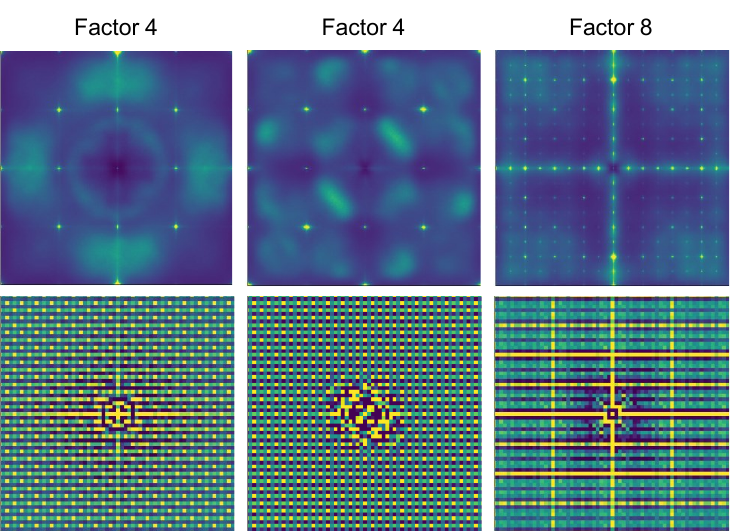}
        \caption{Power spectra (top) and autocorrelation functions (bottom) of noise residuals of three slightly different latent diffusion models. The upsampling factor between latent space and pixel space is 4 for the first two architectures and 8 for the third one. This single parameter appears to be responsible for the positioning of the  peaks in the power spectra and the periodicity of the regular patterns in the autocorrelations.}
	\label{fig:architecture}
\end{figure}

It is well known that many GAN-based generators leave clear traces of their processing pipeline in the images.
These can be exposed by extracting and averaging the noise residuals, to compute a model artificial fingerprint,
and analyzing the results in the spatial and/or spectral domain.
Images generated by diffusion models appear to show artifacts of a similar nature and, arguably, a similar origin.
Fig.\ref{fig:architecture} shows the power spectra (top) and autocorrelation functions (bottom)
of the artificial fingerprints of three slightly different latent diffusion architectures.
The first two models are characterized by an upsampling factor of 4 between latent space and pixel space and,
not by chance,
exhibit clear peaks in the spectrum at frequencies multiples of 1/4,
and strong regular patterns with periodicity 4 in the autocorrelation function.
The minor differences between the two are mostly due to the different training set used, ImageNet vs. CelebHQ.
On the contrary, the third model, trained on LSUN churches images, has an upsampling factor of 8.
As expected, peaks in the power spectrum are now denser, at frequencies multiple of 1/8,
while they are sparser in the autocorrelation function.

While these artifacts are clearly present in current generation architectures,
and can be exploited for forensic needs,
these analyses refer to images observed ``in the lab'', as they leave the generator, with no further processing.
This is not a realistic condition,
just think of all the standard processing steps, especially compression and resizing,
applied to images as soon as they are uploaded on a social network.
These post-processing steps may significantly modify and hide the artifacts highlighted before
and should be always taken into account when designing a forensic detector.

\begin{figure}[t!]
        \centering
	\includegraphics[width=1.0\linewidth,clip,trim=183 10 183 0,page=6]{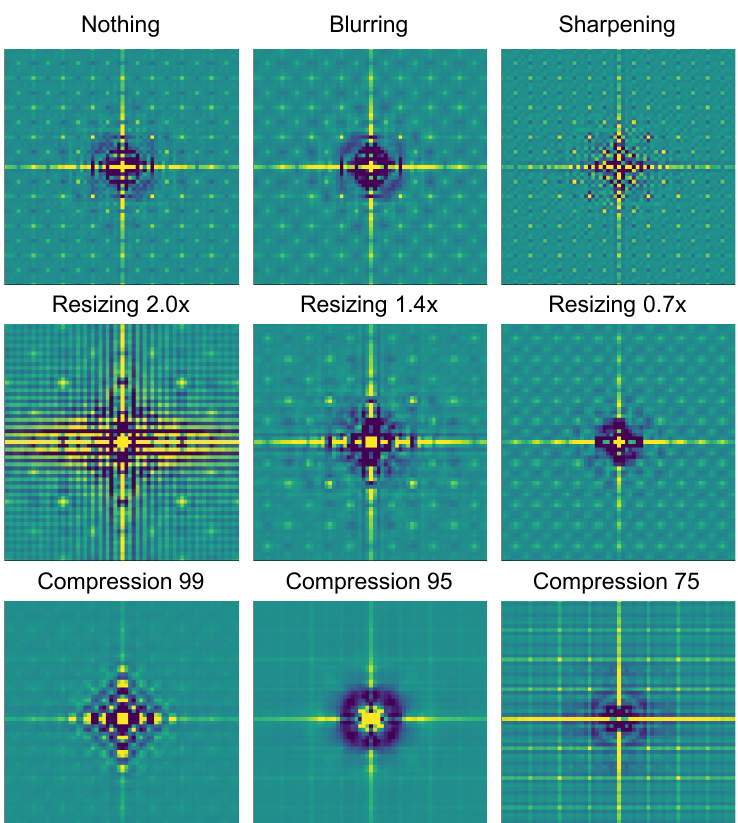}
        \caption{Effect of post-processing operations on the autocorrelation of noise residuals. We consider Stable Diffusion images and apply to them different operations to see how the autocorrelation changes. In some cases the original generation traces (top-left) are completely hidden by post-processing artifacts.}
	\label{fig:processing}
\end{figure}

Fig.\ref{fig:processing} shows some interesting examples.
We consider images generated by Stable Diffusion and show, on the top-left corner,
the autocorrelation function observed in the absence of any post-processing.
A clear regular pattern is visible
which suggests the artificial nature of these images and may even be used to trace back the individual generator used.
In the other slots, we show the autocorrelation function after various types of post-processing,
blurring, sharpening, resizing at three different scales, JPEG compression at three different quality factors.
The autocorrelation function changes more or less markedly depending on the nature and  intensity of the post-processing, but in most cases it can be hardly associated with the original one.
In the presence of strong JPEG compression, for example, the compression artifacts dominate the scene and hide completely the generation artifacts.

\subsection{Training dataset bias}

\begin{figure}[t!]
        \centering
        \includegraphics[width=1.0\linewidth,clip,trim=125 132 241 0,page=3]{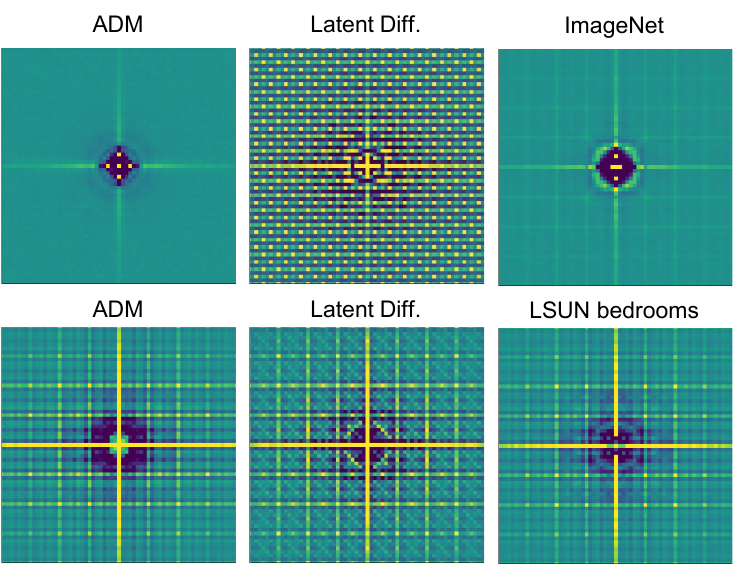}
        \caption{Autocorrelation of noise residuals for real and synthetic images. 
        Top: two diffusion models (ADM and Latent) and the dataset used for their training (ImageNet).
        Bottom: two diffusion models (ADM and Latent) and the dataset used for their training (LSUN Bedroom). 
        In the first case the JPEG traces ($8\times 8$ grid) are barely visible in the training dataset and they are not present in the generated images.
        Instead, in the second case they are clearly visible both in the training dataset and in the generated images.
         }
	\label{fig:compression}
\end{figure}

In our analysis, we singled out also a further problem that undermines the reliability of high-frequency fingerprints. This is exemplified by the images of Fig.~\ref{fig:compression} relative to the DM models, where noise residuals exhibit an obvious regular pattern that may suggest their artificial origin. However, this is the very same pattern observed in the residuals of real images JPEG compressed at quality factor Q=75 (see Fig.~\ref{fig:processing}).
Analyzing in more detail the generation pipeline, we discovered that the dataset used to train the model was composed entirely by images JPEG compressed at Q=75. The model learned to include the compression artifacts in the generated images, that is, it transferred dataset-related artifacts in the generated image with the potential to trick fingerprint-based detectors. This example rings also a bell about the experimental protocol. In fact, most of the times we do not know the details of the dataset used to train a model, but such details may well impact the characteristics of the generated images. A fully reliable analysis must keep this aspect under control.

Interestingly, the ability of GANs models to transfer the same fingerprints from training data to the generated images has been exploited to develop an approach to protect synthetic images by embedding the same artificial fingerprint as proposed in \cite{Yu2021}.

\begin{figure*}[t!]
       \centering
       \includegraphics[width=0.32\linewidth, page=1, clip, trim=0 0 0 0]{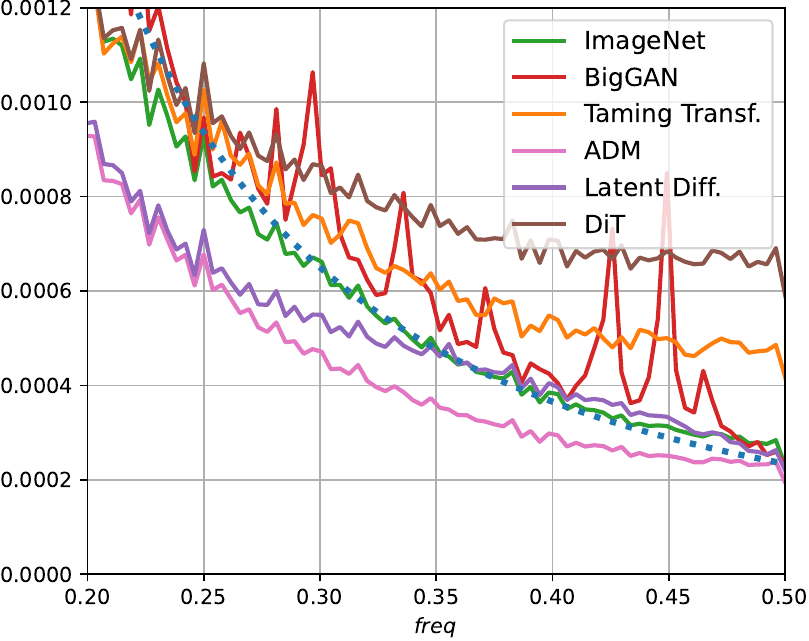}
       \includegraphics[width=0.32\linewidth, page=2, clip, trim=0 0 0 0]{spectrum_radius_imgs_200.pdf}
       \includegraphics[width=0.32\linewidth, page=3, clip, trim=0 0 0 0]{spectrum_radius_imgs_200.pdf}
        \caption{Radial spectrum power density. Synthetic images are compared with the real images used for training the correspondent model. Real images (green) fit very well the expected theoretical curve (dotted).}
 \label{fig:radial_spectrum}
\end{figure*}

\subsection{Fidelity to natural images}

Natural images are known to exhibit strong statistical regularities in the frequency domain.
In particular, several studies \cite{Burton1987, Tolhurst1992}
observed that the average power spectrum decades with a $1/f^\alpha$ law, with $\alpha \simeq 2$.
Moreover, it has been also recognized \cite{Switkes1978, Torralba2003statistics} that the power spectrum is far from isotropic,
with power concentrated especially at the horizontal and vertical orientations and much less at the others.
These may be precious pieces of information towards the detection of synthetic images.
Indeed, considering their good photorealism and high semantic quality,
we expect their spectra to be very similar to those of real images, at low frequencies.
However, it is certainly possible that significant deviations arise at mid-high frequencies,
not because of generation artifacts but due to the inability to correctly replicate the statistics of natural images.
This conjecture is indirectly supported also by the literature
where there is wide agreement on the superior performance of dual-stream detectors,
which exploit both high-frequency artifacts and original RGB images.
With this approach, the detector has the opportunity to analyze not only the highest frequencies, with their subtle generation artifacts,
but also the intermediate frequency, which carry information on the signal itself,
with beneficial effects.

\vspace{2mm} \noindent
\bb{Analysis of radial spectra.}
To test this conjecture, we now analyze the radial spectra of real and synthetic images.
The radial spectrum of the $i$-th image is computed
by averaging the values of its Fourier transform (normalized by its standard deviation) at distance $\rho$ from the origin, irrespective of the angular direction
\begin{equation}
    X_i(\rho) = \langle X_i(k,l) \rangle_{A(\rho)}
\end{equation}
where $A(\rho)$ is the annulus between radii $\rho-\Delta\rho$ and $\rho+\Delta\rho$, and $\rho$ goes from 0 to 0.5 in 128 discrete steps.
Then, as usual, an average on $I=1000$ images is taken to characterize the whole source.
In our analyses we will discard low and very high frequencies and focus on the range $\rho \in [0.2-0.5]$.

Our goal here is to test whether the radial spectra of synthetic images follow the same ``correct'' law as those of real images.
However, to avoid dataset-related biases,
for a given real source ({\it e.g.}, imagenet), we study only the images that are generated using the very same source as training set.
With this care, statistical difference can be attributed only to the generation engine.
In Fig.~\ref{fig:radial_spectrum}, for a few widespread real sources, we compare radial spectra of real and corresponding generated images.
The former (green) decay with a nearly perfect power law (dashed black).
On the contrary, all radial spectra of synthetic sources deviate more or less markedly and in one or more frequency ranges from that law.
Generally, but not always, real images seem to possess a richer content at the intermediate frequencies.
Only for the FFHQ dataset, the radial spectra of some synthetic sources approximate pretty well that of the real source.
However, the FFHQ dataset is composed by images that have been interpolated, that is low-pass filtered,
a process that inevitably modifies the original spectral content and makes it indistinguishable from that of some synthetic images.
In hindsight, this further supports our conjecture.

\begin{figure*}[t!]
       \centering
       \setlength{\tabcolsep}{0pt}
       \renewcommand\arraystretch{0.0}
       \begin{tabular}{ccc}
       \includegraphics[width=0.32\linewidth, page=1, trim=0 -15 0 0]{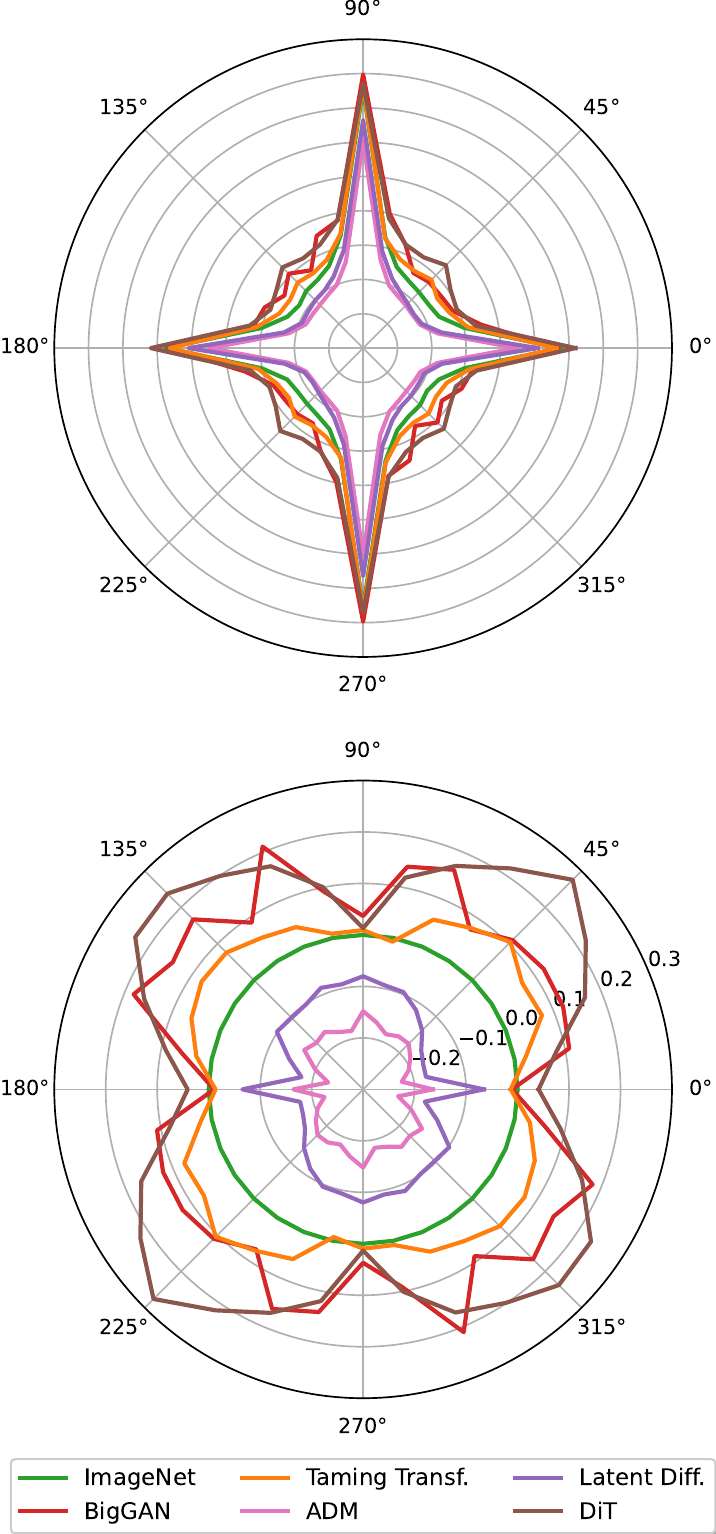} &
       \includegraphics[width=0.32\linewidth, page=2]{spectrum_angular_imgs_200.pdf} &
       \includegraphics[width=0.32\linewidth, page=3]{spectrum_angular_imgs_200.pdf}
       \end{tabular}
        \caption{Angular spectrum power density (top) and the square-root of Fisher's discriminant ratio applied at each direction (bottom).}
	\label{fig:angular_spectrum}
\end{figure*}

\vspace{2mm} \noindent
\bb{Analysis of angular spectra.}
A similar analysis is carried out for the angular spectra, where now
\begin{equation}
    X_i(\theta) = \langle X^{hp}_i(k,l) \rangle_{C(\theta)}
\end{equation}
$C(\theta)$ is a double cone between angles $\theta-\Delta\theta$ and $\theta+\Delta\theta$, and $\theta$ goes from 0 to $\pi$ in 16 discrete steps.
Note that the non-informative low-pass content is filtered beforehand with cut-off frequency 0.1.
The resulting plots are reported in Fig.~\ref{fig:angular_spectrum}(top) for the same real and synthetic sources as before.
Mostly, the very same considerations already done for radial spectra apply.
The only exceptions are observed again for the FFHQ dataset and for the same reasons as before.
In this case, there is no theoretical reference pattern from the literature.
We observe again significant deviations from the angular spectra of the real sources (green).
Such deviations can be better appreciated in the plots of Fig.~\ref{fig:angular_spectrum}(bottom), showing the square root of the Fisher's discriminant ratios
\begin{equation}
    F_s(\theta) = \frac{\mu_s(\theta)-\mu_0(\theta)}{\sqrt{\sigma^2_s(\theta)+\sigma^2_0(\theta)}}
\end{equation}
where $\mu_s(\theta)=X_s(\theta)$ and $\mu_0(\theta)=X_0(\theta)$ are the mean angular spectra of synthetic ($s$) and real ($0$) images, respectively,
and $\sigma^2_s(\theta), \sigma^2_0(\theta)$ the corresponding variances.
In this case the real source is represented by the green unit circle.
Finally, it is worth noting that the StyleGAN2 and StyleGAN3 models better fit the spectral distribution of real images.
For StyleGAN3, this is justified by the development of GAN architectures that were able to avoid aliasing\cite{karras2021alias}.

\section{Conclusions}

Generators of synthetic images are becoming more and more sophisticated.
Casual viewers are already unable to tell real images from fake ones
and even forensic tools struggle to cope with the rapid progress of new models.
In this context, brute-force detectors, even trained on huge amounts of data, may not be up to the task,
especially when asked to generalize to new generators.
Therefore, it is important to single out and possibly exploit all features that distinguish real from fake data.
In this work, we analyzed a large number of modern and popular generators
looking for such features both in the spatial domain and in the frequency domain.
In both domains we relied on second-order statistics, autocorrelation functions and power spectra.

This investigation provides some interesting results.
On one hand, even the most sophisticated architectures keep generating artifacts that can be exploited for detection.
These may be more visible in the spatial or in the spectral domain and both should be always explored.
In addition, generators do not seem able to mimic the spectral distribution of real images at the mid-high frequencies
and radial and angular spectra may help expose such discrepancies.
Finally, we noted that a strongly biased training set may teach the model
to transfer the very same biases in the generated images,
a possible problem for both generators and detectors.
In future work we will pursue the design of forensic detectors based also on these traces.

\paragraph{Acknowledgment.}
We gratefully acknowledge the support of this research by the Defense Advanced Research Projects Agency (DARPA) under agreement number FA8750-20-2-1004.
The U.S. Government is authorized to reproduce and distribute reprints for Governmental purposes notwithstanding any copyright notation thereon.
The views and conclusions contained herein are those of the authors and should not be interpreted as necessarily representing the official policies or endorsements, either expressed or implied, of DARPA or the U.S. Government.
This work has also received funding by the European Union under the Horizon Europe vera.ai project, Grant Agreement number 101070093, and is supported by a TUM-IAS Hans Fischer Senior Fellowship and by the PREMIER project funded by the Italian Ministry of Education, University, and Research within the PRIN 2017 program.
Finally, we thank Tero Karras, Yogesh Balaji, and Ming-Yu Liu for sharing data for StyleGAN-T and Ediffi-I experiments.

\bibliographystyle{ieee_fullname}
{\small \bibliography{refs}}
\end{document}